\def\year{2017}\relax
\documentclass[letterpaper]{article}
\usepackage{aaai17}
\usepackage{times}
\usepackage{helvet}
\usepackage{courier}

\usepackage{graphicx}

\usepackage{bigstrut}
\usepackage{booktabs}
\usepackage[utf8]{inputenc} 
\usepackage[T1]{fontenc}    
\usepackage{url}            
\usepackage{booktabs}       
\usepackage{amsfonts}       
\usepackage{nicefrac}       
\usepackage{microtype}      
\usepackage{subfigure}
\usepackage{amsmath}
\usepackage{mathtools}
\usepackage{caption}

\begin{document}
%
\title{Accurate Deep Representation Quantization with Gradient Snapping Layer for Similarity Search}
\author{Shicong Liu, Hongtao Lu
}
\maketitle

\begin{abstract}
Recent advance of large scale similarity search involves using deeply learned representations to improve the search accuracy and use vector quantization methods to increase the search speed. However, how to learn deep representations that strongly preserve similarities between data pairs and can be accurately quantized via vector quantization remains a challenging task. Existing methods simply leverage quantization loss and similarity loss, which result in unexpectedly biased back-propagating gradients and affect the search performances. To this end, we propose a novel gradient snapping layer (GSL) to directly regularize the back-propagating gradient towards a neighboring codeword, the generated gradients are un-biased for reducing similarity loss and also propel the learned representations to be accurately quantized. Joint deep representation and vector quantization learning can be easily performed by alternatively optimize the quantization codebook and the deep neural network. The proposed framework is compatible with various existing vector quantization approaches. Experimental results demonstrate that the proposed framework is effective, flexible and outperforms the state-of-the-art large scale similarity search methods.
	
\end{abstract}

\section{Introduction}
Approaches to large scale similarity search typically involves compressing a high-dimensional representation into a short code of a few bytes by hashing \cite{SH,itq} or vector quantization \cite{pq}. Short code embedding greatly save space and accelerate the search speed. A trending research topic is to generate short codes that preserves semantic similarities, e.g., \cite{lin2014fast,sq}. Traditional short code generation methods require that the input data is firstly represented by some hand-crafted representations (e.g., GIST \cite{oliva2001modeling} for images). Recently, learned representations with a deep neural network and a similarity loss is more preferable due to its stronger ability to distinguish semantic similarities\cite{chopra2005learning,wang2014learning}.

In this paper we explore learning effective deep representations and short codes for fast and accurate similarity search. \cite{xia2014supervised,lai2015simultaneous} propose to learn deep representations and mappings to binary hash codes with a deep convolutional neural network. However these deeply learned representations must exhibit certain clustering structures for effective short codes generation or severe performance loss could induce. How to learn representations that can be converted into short codes with minimal performance loss still remains a challenging task. Existing methods \cite{zhao2015deep,zhu2016deep,cao2016deep} add a quantization loss term to the similarity loss to regularize the output to be \textit{quantizable}. However this methodology leads to difficulties in optimization, since they actually introduce a constant bias in the gradients computed from similarity loss for each sample. A joint representation and quantization codebook learning framework which is easy to be optimized and is compatible with various quantization methods is thus highly desirable. 

\begin{figure*}[t]
	\centering
	\includegraphics[width=0.9\linewidth]{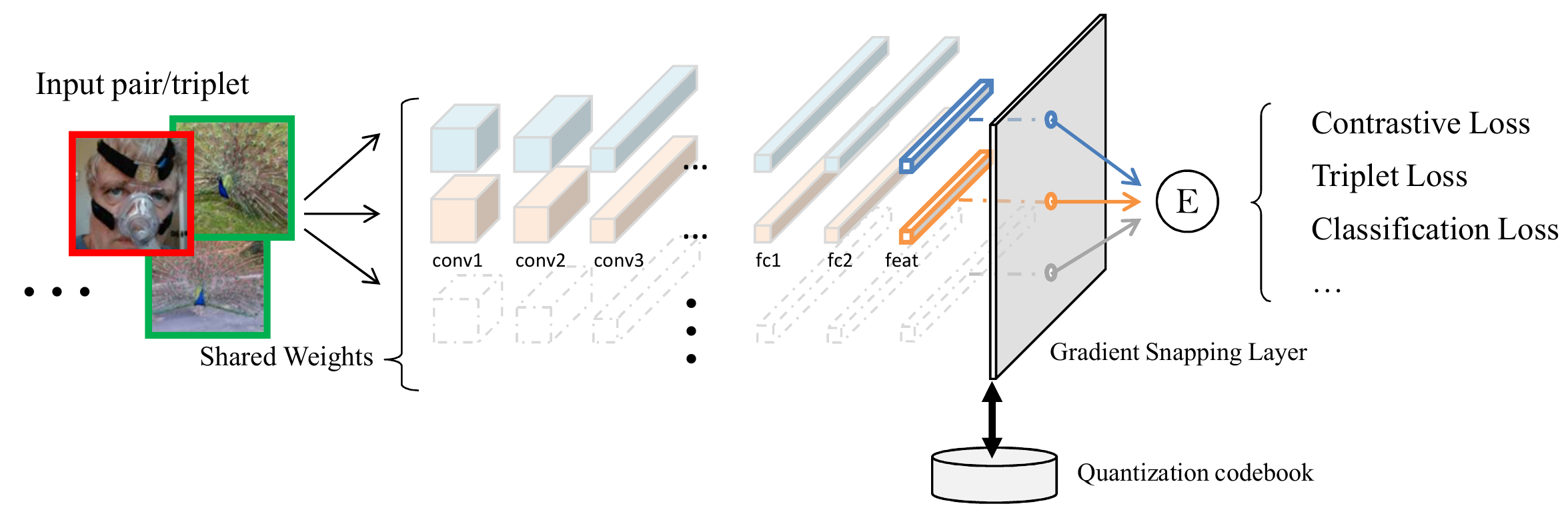}
	\caption{Illustrative network configuration with gradient snapping layer for deep representation quantization. We use a siamese network architecture, which takes several input images and computes network output with the same weights (also named shared weights). A gradient snapping layer associating a quantization codebook is inserted before the loss function, which could be any of the contrastive loss, triplet loss or other commonly used loss functions for neural network.}
	\label{illuNet}
\end{figure*}

Our main contributions are as follows: \textbf{(i)} We put forward a novel \textit{gradient snapping layer} (GSL) to learn representations exhibiting required structures for vector quantization: Instead of directly regularizing the network output by adding a quantization loss, we regularize the back-propagating gradients by \textit{snapping} the gradient computed from similarity loss toward an optimal neighbor codeword. By doing so we still reduce the quantization error and now the computed gradients are un-biased. \textbf{ (ii)} GSL is yet another layer within existing deep learning framework and is compatible with any network structure. An illustrative configuration is presented in Fig.\ref{illuNet}. \textbf{(iii)} Our framework is compatible with any state-of-the-art vector quantization methods. GSL regularizes the output of its top layer to be \textit{quantizable} by arbitrary vector quantization codebook. \textbf{(iv)} Experimental results on ILSVRC 2012 \cite{deng2009imagenet}, CIFAR \cite{krizhevsky2009learning}, MNIST \cite{lecun1998gradient} and NUS-WIDE \cite{chua2009nus} show that our joint optimization framework significantly outperforms state-of-the-art methods by a large margin in term of search accuracy.

\section{Related Works}

\subsection{Deep Representations for Similarity Search}

Convolutional Neural Networks have shown strong performances in image classification tasks. Qualitative \cite{krizhevsky2012imagenet} and quantitative \cite{babenko2014neural} evidences of their feasibility for image retrieval tasks have also been observed in the former studies. Convolutional networks can generate effective deep representations within the \textit{siamese architectures}\cite{chopra2005learning}, and are successful in fine-grained classification \cite{wang2014learning}, unsupervised learning\cite{wang2015unsupervised}, and image retrieval\cite{xia2014supervised,lai2015simultaneous,zhao2015deep,cao2016deep} tasks, etc. A Siamese architecture consists of several streams of layers sharing the same weights and uses a multi-wise loss function to train the network.  
A commonly used loss function for retrieval tasks is triplet loss\cite{lai2015simultaneous}, since it characterizes a relative ranking order and achieves satisfying performances\cite{kumar2015learning}. Formally, given a sample $\mathbf{x}$, which is more similar to $\mathbf{x}^+$ rather than $\mathbf{x}^-$, the triplet loss is defined as :
\begin{equation}
E=\textit{max} \{0, g+\lVert\hat{\mathbf{y}}_{\mathbf{x}} - \hat{\mathbf{y}}_{\mathbf{x}^+}\rVert_2 - \lVert\hat{\mathbf{y}}_{\mathbf{x}} - \hat{\mathbf{y}}_{\mathbf{x}_-}\rVert_2\}
\label{Etrip}
\end{equation}

, where $\hat{\mathbf{y}}_\mathbf{x}$, $\hat{\mathbf{y}}_\mathbf{x^-}$, $\hat{\mathbf{y}}_\mathbf{x^+}$ are the deep representations of $\mathbf{x}, \mathbf{x}_-, \mathbf{x}_+$, respectively. $g$ is a gap parameter that regularizes the distance difference between "more similar" pair and "less similar" pair, it also prevents network from collapsing. Pair-wise loss \cite{cao2016deep} and classification loss are also used in recent literatures \cite{sq} due to their simplicity for optimization.

Compared to traditional hand-crafted visual features, deeply learned representations significantly improves the performances of image retrieval. However the deep representations obtained with convolutional neural networks are high dimensional (e.g., 4096-d for AlexNet \cite{krizhevsky2012imagenet}), thus the representations must be effectively compressed for efficient retrieval tasks.

\subsection{Vector Quantization for Nearest Neighbor Search}
Vector quantization methods are promising for nearest neighbor retrieval tasks. They outperform the hashing competitors like ITQ\cite{itq}, Spectral Hashing\cite{SH}, etc, by a large margin\cite{pq,opq} on metrics like $l_2$. Vector quantization techniques involve quantizing a vector as a sum of certain pre-trained vectors and then compress this vector into a short encoding representation. Taking Product Quantization (PQ) \cite{pq} as an example, denote a \emph{database} $\mathcal{X}$ as a set of $N$ $d$-dimensional
vectors to compress, PQ learns $M$ \emph{codebooks} $\mathcal{C}_{1}, \mathcal{C}_{2}, \cdots, \mathcal{C}_{M}$ for disjoint $M$ subspaces, each of which is a list of $K$ \textit{codewords}: $\mathbf{c}_m(1)$, $\mathbf{c}_m(2)$, $\ldots$, $\mathbf{c}_m(K) \in \mathbb{R}^{d/M}$. Then PQ uses a \emph{mapping function} $i_m(\cdot)\,:\,\mathbb{R}^{d/M}\,\to\,[K]$\footnote{$[K]$ denotes $\{1,\,2,\,3,\,\ldots,\,K\}$} to encode a vector: $\mathbf{x}\mapsto (i_1(\mathbf{x}), \cdots, i_M(\mathbf{x}))$. A \emph{Quantizer} $q_m$ is defined as $q_m(\mathbf{x})=\mathbf{c}_m(i(\mathbf{x}))$,
meaning the $m$-th subspace of $\mathbf{x}$ is approximated as $q(\mathbf{x})$ for latter use. Product quantization minimizes the \emph{quantization error}, which is defined as
\begin{equation}
E=\frac{1}{N}\sum_{\mathbf{x}\in\mathcal{X}} ||\mathbf{x}-[q_1(\mathbf{x}), \cdots, q_M(\mathbf{x})]^T||^{2}.\label{E}
\end{equation}
Minimizing Eqn.\ref{E} involves performing classical K-means clustering algorithm on each of the $M$ subspaces. Afterwards, one can use \emph{asymmetric distance computation} (ADC) \cite{pq} to greatly accelerate distance computation: Given a query vector $\mathbf{q}$, we first split it to $M$ subspaces: $\mathbf{q}=[\mathbf{q}_{1}^{\top},\,\cdots,\,\mathbf{q}_{M}^{\top}]^{\top}$. We then compute and store $\lVert\mathbf{q}_{m}-\mathbf{c}_{m}(k)\rVert^{2}$
for $k\in[K]$ and $m\in[M]$ in a look-up table. Finally, for each $\mathbf{x}\in\mathcal{X}$, the distance between
$\mathbf{q}$ and $\mathbf{x}$ can be efficiently approximated using only $(M-1)$
floating-point additions and $M$ table-lookups with the following equation:
\begin{equation}
\lVert\mathbf{q}-\mathbf{x}\rVert^{2}\approx\sum_{m=1}^{M}\lVert\mathbf{q}_{m}-\mathbf{c}_{m}(i_{m}(\mathbf{x}))\rVert^{2}
\end{equation}
A number of improvements for PQ have been put forward, e.g., \emph{Optimized
	product quantization} (OPQ) \cite{opq} \emph{additive quantization} (AQ)
\cite{babenko2014additive}, and \textit{generalized residual vector quantizaiton} (GRVQ) \cite{grvq}, etc. Vector quantization methods also allow a number of efficient non-exhaustive search methods including Inverted File System\cite{pq}, Inverted Multi-index\cite{babenko2012inverted}.

Recent studies on vector quantization focus on vector quantization for semantic similarities, for example, supervised quantization\cite{sq} applies a linear transformation that discriminates samples from different classes on the representations before applying composite quantization\cite{composite}. This approach is still upper-bounded by the input representations.

\subsection{Joint Representation and Quantization Codebook Learning}
To fully utilize deep representations for similarity search, the learned representation must strongly preserve similarities, and exhibit certain clustering structures to be \textit{quantizable} for effective short code generation. This is a typical multi-task learning problem. Existing methods including deep quantization network (DQN) \cite{cao2016deep}, deep hashing network (DHN) \cite{zhu2016deep} , simultaneous feature and hashing coding learning (SHFC)\cite{lai2015simultaneous}, etc, all propose to  leverage between similarity loss $E(\cdot)$  and quantization loss $Q(\mathbf{y})=\lVert \mathbf{y}-q(\mathbf{y})\rVert^2$ , and derive the final loss function\footnote{The methodology is the same though they could adopt different similarity/quantization loss function.}:
\begin{equation}
E'(\mathbf{y})=E(\mathbf{y})+\lambda Q(\mathbf{y})
\label{YYY}
\end{equation}
However, the above loss function is actually not suitable for joint representation and quantization codebook learning. Consider the following:
\begin{equation}
\begin{split}
\frac{\partial E'(\mathbf{y})}{\partial \mathbf{y}}&=\frac{\partial E(\mathbf{y})}{\partial \mathbf{y}}+\lambda \frac{\partial Q(\mathbf{y})}{\partial \mathbf{y}} \\
&=\frac{\partial E(\mathbf{y})}{\partial \mathbf{y}}+2\lambda (\mathbf{y}-q(\mathbf{y}))\\
&=\frac{\partial E(\mathbf{y})}{\partial \mathbf{y}}+2\lambda \epsilon_\mathbf{y}
\end{split}
\end{equation}
where $\epsilon_\mathbf{y}$ denotes the quantization residual vector of $\mathbf{y}$. Thus existing methods can be considered as simply adding a bias term specific to each sample. This is not desirable since we want to find a gradient that both reduce the similarity loss and the quantization loss, the biased similarity loss distorts the gradients computed from similarity loss and eventually leads to sub-optimal representations.

\begin{figure}[t]
	\centering
	\includegraphics[width=1\linewidth]{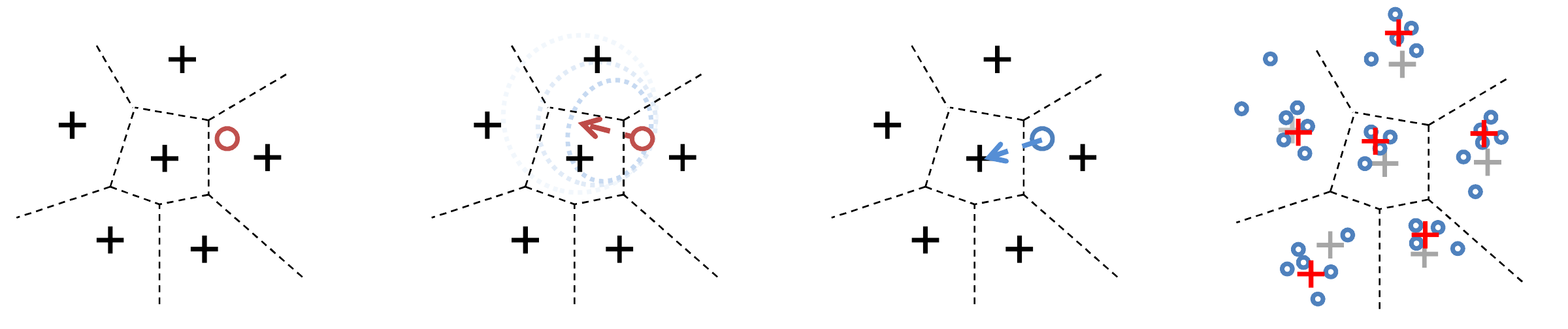}
	\caption{Illustrative demonstration of training with GSL. From left to right: 1) Red circle: raw representations; black crosses: codewords. 2) Compute the gradient from similarity loss, then enumerate neighboring codewords for gradient snapping. 3) Snap the gradient with Eqn.\ref{gequ}. 4) Updated $\mathcal{C}$.}
	\label{illuLayer}
\end{figure}
\interfootnotelinepenalty=10000

\begin{figure*}[t]
	\centering
	\subfigure[]{\includegraphics[width=0.400\linewidth]{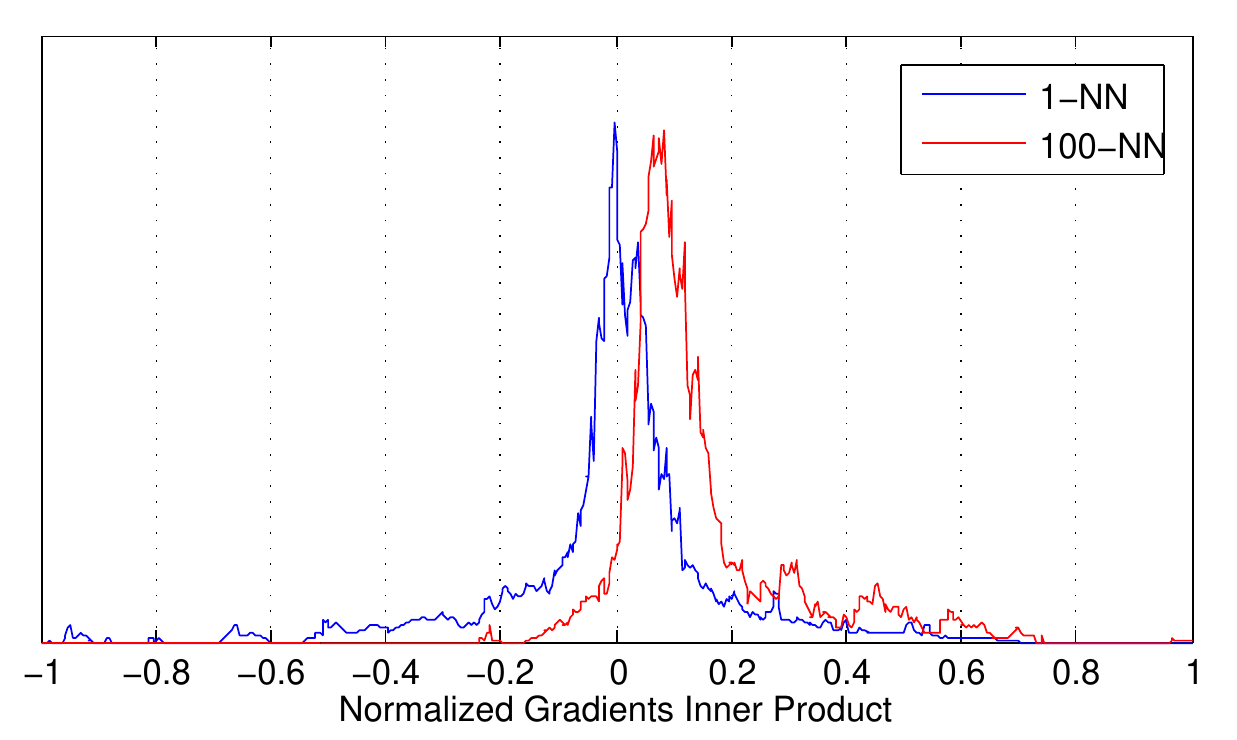}\label{sal:ovg}} 
	\subfigure[]{\includegraphics[width=0.263\linewidth]{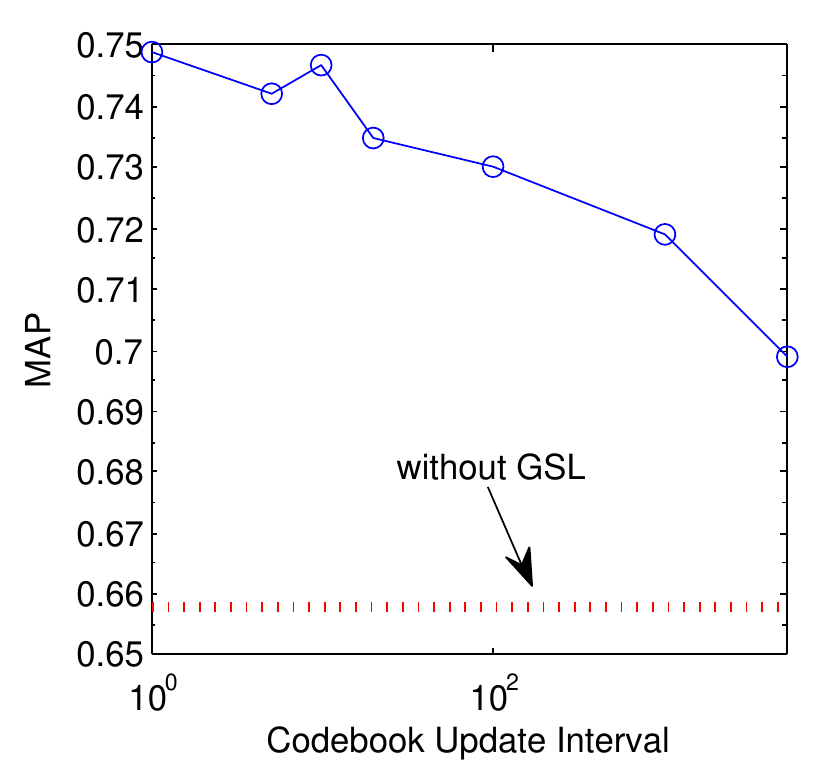}\label{sal:ofq}} 
	\subfigure[]{\includegraphics[width=0.245\linewidth]{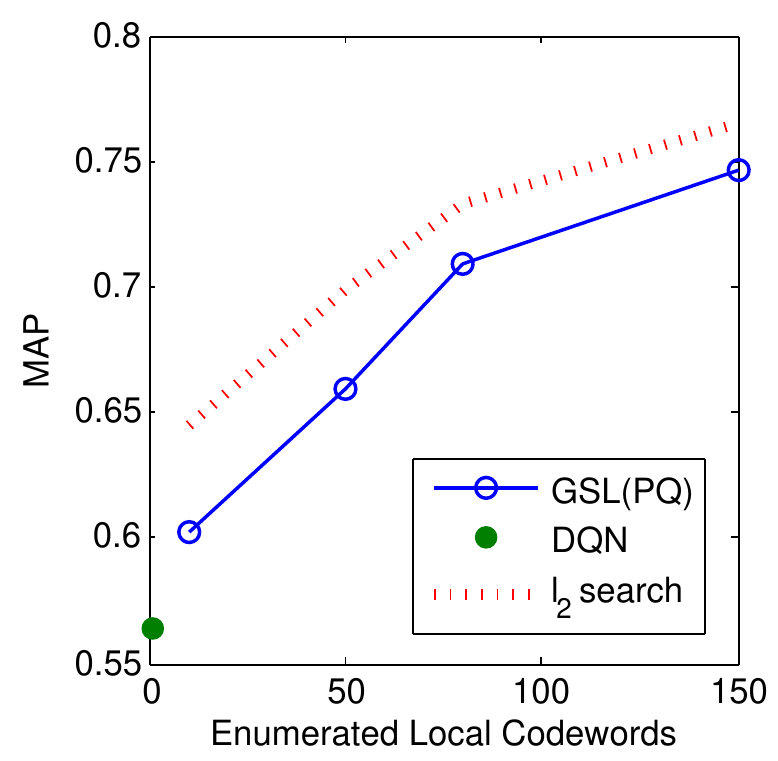}\label{sal:nn}} 
	\caption{ Empirical analysis on CIFAR10, using triplet loss, and associate with 32-bit product quantization codebook. a) Comparison between output regularization and gradient snapping. We normalize $\partial E/\partial \mathbf{y}$ and $\Delta \mathbf{c}$, then draw the histogram of their inner products. b) The impact of codebook update interval on retrieval performance measured in MAP (mean average precision), with 150 enumerated codewords. c) How the number of enumerated neighboring codewords impacts retrieval performance, with codebook update interval 1.
	}
	\label{sal} 
\end{figure*}

\newcommand{\specialcell}[2][c]{%
	\begin{tabular}[#1]{@{}c@{}}#2\end{tabular}}
\setlength{\tabcolsep}{0.4em}
\begin{table*}[t]
	\small
	\caption{Image retrieval performances in MAP of 32-bit code obtained with different representations on CIFAR-10. Value in brackets are reported from \cite{lai2015simultaneous,shen2015supervised}; values with * are trained on all 59K training samples.}
	\centering
	\begin{tabular}{cccccccccc}
		\toprule
		& $l_2$ & PQ    & FashHash & SDH   & ITQ   & AQ& OPQ &KSH& \\ \midrule
		GIST         & \textit{0.177} & 0.170 &\specialcell{ 0.372\\ (\textasciitilde  0.37*)}    & \specialcell{\textbf{0.400} \\(\textasciitilde  0.43*)} & \specialcell{0.174 \\ (0.172)} & 0.175& 0.178 &(0.346)& \\
		AlexNet 4096-d    & \textit{0.313} & 0.264 & \textbf{0.603}    & 0.586 & 0.260 & 0.233& 0.292& (0.558) & \\
		\textit{feat}, no GSL   & \textit{0.779} & 0.658 & 0.646    & 0.619 & 0.582 & 0.669 & \textbf{0.683} & - & \\
		\textit{feat}+GSL(PQ) & \textit{0.765} & \textbf{0.747} & 0.641    & 0.624 & 0.603 & 0.691& -&-& \\
		\textit{feat}+GSL(AQ) & \textit{0.751} & 0.639 & 0.638    & 0.615 & 0.590 & \textbf{0.712} & -&-& \\ \bottomrule& 
	\end{tabular} 
		
	\label{tabmvvv}
\end{table*}
\section{Gradient Snapping Layer}
We propose \textit{gradient snapping layer} (GSL) to jointly learn representations and to enforce the learned representations to exhibit cluster structure. We first start from the following loss function leveraging similarity loss and an adaptive quantization loss:
\begin{equation}
\begin{split}
& E'(\mathbf{y})=\lambda_1 E(\mathbf{y})+\lambda_2 \lVert \mathbf{y}-\mathbf{c} \rVert^2, \\
&\text{s.t. } \text{arg}\max_{\mathbf{c}\in\mathcal{C}} (\frac{\partial E(\mathbf{y})}{\partial \mathbf{y}} )^T \frac{\partial E'(\mathbf{y})}{\partial \mathbf{y}} \cdot f(\lVert \mathbf{c}-\mathbf{y} \rVert^2)
\end{split}
\label{XXX}
\end{equation}
$f(\cdot) \in \mathbb{R}^+$ is a function that leverages between different neighboring codewords of $\mathbf{y}$. A neighboring codeword is more preferred compared to distant codeword. We may choose a Gaussian function $f\left(d\right) = e^{- d^2/\sigma } $ where $\sigma$ is the mean $l_2$ distance between $\mathbf{y}$ and the neighboring codewords. It's easy to see that the above loss function is unbiased by introducing a small variance to the quantization loss term. Eqn.\ref{YYY} is a special case when we choose $\mathbf{c}=q(\mathbf{y})$. 

Based on Eqn.\ref{XXX}, we can learn \textit{quantizable} and similarity preserving representations by GSL by inserting it between the representation layer and the similarity loss layer. It works like snapping the back-propagating gradients towards a codeword: On the forward-pass phase, it directly outputs the representation vector $\mathbf{y}$. On the backward-pass phase, it takes gradient $\frac{\partial E}{\partial\mathbf{y}}$ from the similarity loss layer, and solves Eqn.\ref{XXX} by enumerating a number of nearest codewords $\mathbf{c}\in\mathcal{C}$ to $\mathbf{y}$ and minimizing the following:
\begin{equation}
\begin{split}
&\arg\max_{\mathbf{c}} (\frac{\partial E}{\partial\mathbf{y}}) ^T \cdot \Delta \mathbf{c}, \quad \\ &\text{where} \quad \Delta\mathbf{c}=f(\lVert\mathbf{c}-\mathbf{y}\rVert^2)\frac{(\mathbf{c}-\mathbf{y})}{\lVert\mathbf{c}-\mathbf{y}\rVert}
\label{eqn}
\end{split}
\end{equation}
Then we compute and back-propagates the snapped gradient $\Delta \mathbf{y}$ to the representation layer. In this paper we choose a simple yet effective strategy that leverages the original gradient and its projection towards codeword $\mathbf{c}$, which is given below:
\begin{equation}
\begin{split}
\Delta \mathbf{y} & =  \lambda_1 \frac{\partial E}{\partial\mathbf{y}} + \lambda_2 \Delta \mathbf{c}, \\
\text{where } & \lambda_1=(1-\frac{((\frac{\partial E}{\partial\mathbf{y}})^T \Delta \mathbf{c})^2}{\lVert\Delta\mathbf{c}\rVert^2\lVert \frac{\partial E}{\partial\mathbf{y}} \rVert})\lambda, \\
& \lambda_2=\frac{(\frac{\partial E}{\partial\mathbf{y}})^T \Delta \mathbf{c}}{\lVert \Delta \mathbf{c} \rVert }              
\end{split}
\label{gequ}
\end{equation}
, where $\lambda_2$ is the projected gradients on $\mathbf{c}$, $\lambda_1$ computes the residual of original gradients by projecting $\lambda_2\Delta \mathbf{c}$ back and further scaled by $\lambda$. 

Simultaneous to representation training, $\mathcal{C}$ should also be updated to fit the new deep representations. A simple method is to gather the new representations and update $\mathcal{C}$ every a number of iterations. Take PQ as an example, one can use sequential k-means clustering on each subspace streamed with network output to update the codebooks. We present an illustrative explanation of how our gradient snapping layer work in Fig.\ref{illuLayer} for better understanding. 

Note GSL can be seen as yet another type of layer within deep learning framework, and the main use is to regularize the learned representations to be \textit{quantizable}. Thus it's naturally compatible with any loss function (e.g., triplet loss, contrastive loss) or any network structures. We present an typical configuration in Figure \ref{illuNet}. The learned representations are then quantized, stored, queried linearly or with systems like IVFADC and inverted-multi index. Readers may refer to \cite{pq,babenko2012inverted,kalantidis2014locally} for details.

\section{Experiments and Discussions}
\label{exp}
We conducted extensive experiments on the following commonly used benchmark datasets to validate the effectiveness of gradient snapping layer (GSL):

\textbf{NUS-WIDE}\cite{chua2009nus}, a commonly used web image dataset for image retrieval. Each image is associated with multiple semantic labels. We follow the settings in \cite{lai2015simultaneous,cao2016deep}, and use the images associated with the 21 most frequent labels and resize the images into $256\times 256$. \footnote{We gather and use 139,633 images for experiment due to a number of original images are no longer available.}

\textbf{CIFAR-10}\cite{krizhevsky2009learning}, a dataset containing 60,000 $32\times 32$ tiny images in 10 classes commonly used for image classification. It's also a popular benchmark dataset for retrieval tasks. The dataset contains 512-d GIST feature vectors for all images. 

\textbf{MNIST}\cite{lecun1998gradient}, a dataset of 70,000 $28\times 28$ greyscale handwritten digits from 0 to 9. Each image is represented by raw pixel values, resulting a 784-d vector.

\textbf{ILSVRC2012}\cite{deng2009imagenet}, a large dataset containing over 1.2 million images of 1,000 categories. It has a large training set of 1,281,167 images, around 1K images per category. We use the provided validation images as the query set, it contains 50,000 images.

Following previous works of \cite{xia2014supervised} \cite{lai2015simultaneous}\cite{cao2016deep}, for MINST, CIFAR-10 and NUS-WIDE, we sample 100 images per class as the test query set, and 500 images per class as training set. We use all training data in ILSVRC2012 to train the network as in \cite{sq} and use the validation dataset as query.

We evaluate image retrieval performance in term of mean average precision (MAP). We use hamming ranking for hashing based methods and asymmetric distance computation for vector quantization to retrieve the image rankings of the dataset. We limit the length of retrieved images to 1500 for ILSVRC2012 as in \cite{sq}, and 5000 for NUS-WIDE as in \cite{xia2014supervised} etc.

\setlength{\tabcolsep}{0.4em}
\begin{table*}[ht]
	\caption{Comparison between different simultaneous representation and short code learning schemes. *CNNH is under a two-step hashing framework. Other methods perform simultaneous representation and short code learning in an end-to-end framework.}
	\centering
	\begin{tabular}{cllll}
		\toprule
		& Loss function        & Regularization            & Search  &  \\ \midrule
		DQN\cite{cao2016deep}     & Pair-wise cosine     & Product quantization & ADC     &  \\
		DHN\cite{zhu2016deep}     & Pair-wise logistic   & $l_1$ quantization   & Hamming &  \\
		CNNH\cite{xia2014supervised}* & Classification       & No                        & Hamming &  \\
		DRSCH\cite{zhang2015bit}   & Triplet \& pair-wise & No                        & Weighted Hamming &  \\
		DSFH\cite{zhao2015deep}    & List-wise             & Zero-mean                 & Hamming &  \\
		SHFC\cite{lai2015simultaneous}    & Triplet              & No                        & Hamming &  \\
		Ours & Any & Gradient snapping & ADC& \\
		\bottomrule
	\end{tabular} 
	\label{tabcomp}
\end{table*}
\subsection{Network Settings}

We implement GSL on the open-source Caffe deep learning framework. We employ the AlexNet architecture\cite{krizhevsky2012imagenet} and use pre-trained weights for convolutional layers \textit{conv1-conv5, fc6, fc7}. We stack an additional 192-d inner product layer \textit{feat}, GSL layer and triplet-loss layer on \textit{fc7}. We choose triplet loss due to it's typical and commonly used. It also preserve similarity in $l_2$ which is compatible with most existing vector quantization methods. We use the activation at \textit{feat} as deep representation. We employed the triplet selection mechanism described in \cite{wang2014learning} for faster convergence. Small projection transformations are applied to the input images for data augmentation, we choose $\lambda=0.036$. We'll make our implementation publicly available to allow further researches.

\subsection{Empirical Analysis}



We perform experiment on CIFAR-10 to investigate the performances of image retrieval using PQ\cite{pq}, AQ\cite{babenko2014additive}, FashHash\cite{lin2014fast}, SDQ\cite{shen2015supervised}, ITQ\cite{itq}, KSH\cite{liu2012supervised} and using exhaustive $l_2$ search with 512-d GIST representation provided by the dataset, 4096-d activation on layer \textit{fc7} of pretrained AlexNet on ImageNet, and AlexNet fine-tuned with GSL associated with PQ/AQ-32 bit codebook for representation compressing. We retrieve 150 nearest codewords and compute the snapped gradient. We choose 3000 anchors for SDH, 4 as tree depth for FashHash, $M=4, K=256$ for PQ/AQ. The results are shown in Table \ref{tabmvvv}. 

\textbf{Effectiveness of end-to-end learning}: The experimental result again validate the importance of end-to-end learning of representation and short-codes as addressed in \cite{zhao2015deep,zhang2015bit,zhu2016deep}. A fine-tuned representation allow further performance improvements.

\setlength{\tabcolsep}{.18em}
\begin{table*}[ht]
	\caption{Mean Average Precision (MAP) for Different Number of Bits on three benchmark datasets.}
	\centering
	\begin{tabular}{crccccrccccrcccc}
		\toprule
		& \phantom{a} &                 \multicolumn{4}{c}{CIFAR}                  & \phantom{a} &     \multicolumn{4}{c}{MNIST}     & \phantom{a} &   \multicolumn{4}{c}{NUS-WIDE}    \\
		\cmidrule{3-6} \cmidrule{8-11} \cmidrule{13-16}
		Method &             & 12-bit &     24-bit     &     32-bit     &     48-bit      &             & 12-bit & 24-bit & 32-bit & 48-bit &             & 12-bit & 24-bit & 32-bit & 48-bit \\ \midrule
		CNNH*                           &             & 0.484  &     0.476      &     0.472      &      0.489      &             & 0.969  & \textbf{0.975}  & 0.971  & 0.975  &             &          0.617    &   0.663    & 0.657 & 0.688    \\
		SHFC                           &             & 0.552  &     0.566      &     0.558      &      0.581      &             &\multicolumn{4}{c}{N/A}    &             &      0.674  &   0.697    &   0.713    &  0.715    \\
		DQN                            &             & 0.554  &     0.558      &     0.564      &      0.580      &             &\multicolumn{4}{c}{N/A}    &             &      0.768  &   0.776    &   0.783    &   0.792    \\
		DHN                            &             & 0.555  &     0.594      &     0.603      &      0.621      &             & \multicolumn{4}{c}{N/A}     &             &   0.708    & 0.735  & 0.748  & 0.758  \\
		DRSCH   &    & - & 0.622& 0.629& 0.631 & & -& 0.974& 0.979& 0.971& & \multicolumn{4}{c}{N/A$^\dagger$}\\
		DSCH    &   &  -& 0.611 & 0.617& 0.618 & & -& 0.967& 0.972& 0.975& & \multicolumn{4}{c}{N/A$^\dagger$}   \\
		Ours                           &             &   -    & \textbf{0.692} & \textbf{0.747} & \textbf{ 0.752} &             &   -    & 0.973  & \textbf{0.980}  & \textbf{0.981}  &             &   -    & \textbf{0.812}  & \textbf{0.839}  & \textbf{0.849}  \\ \bottomrule
	\end{tabular}
	\label{tabr}
	
	\hfill *An enhanced variant of CNNH proposed in \cite{lai2015simultaneous}
	
	\hfill$^\dagger$ \cite{zhang2015bit} uses a different experimental setting.
\end{table*}
\begin{figure*}[t]
	\centering
	\subfigure[CIFAR10]{\includegraphics[width=0.360\linewidth]{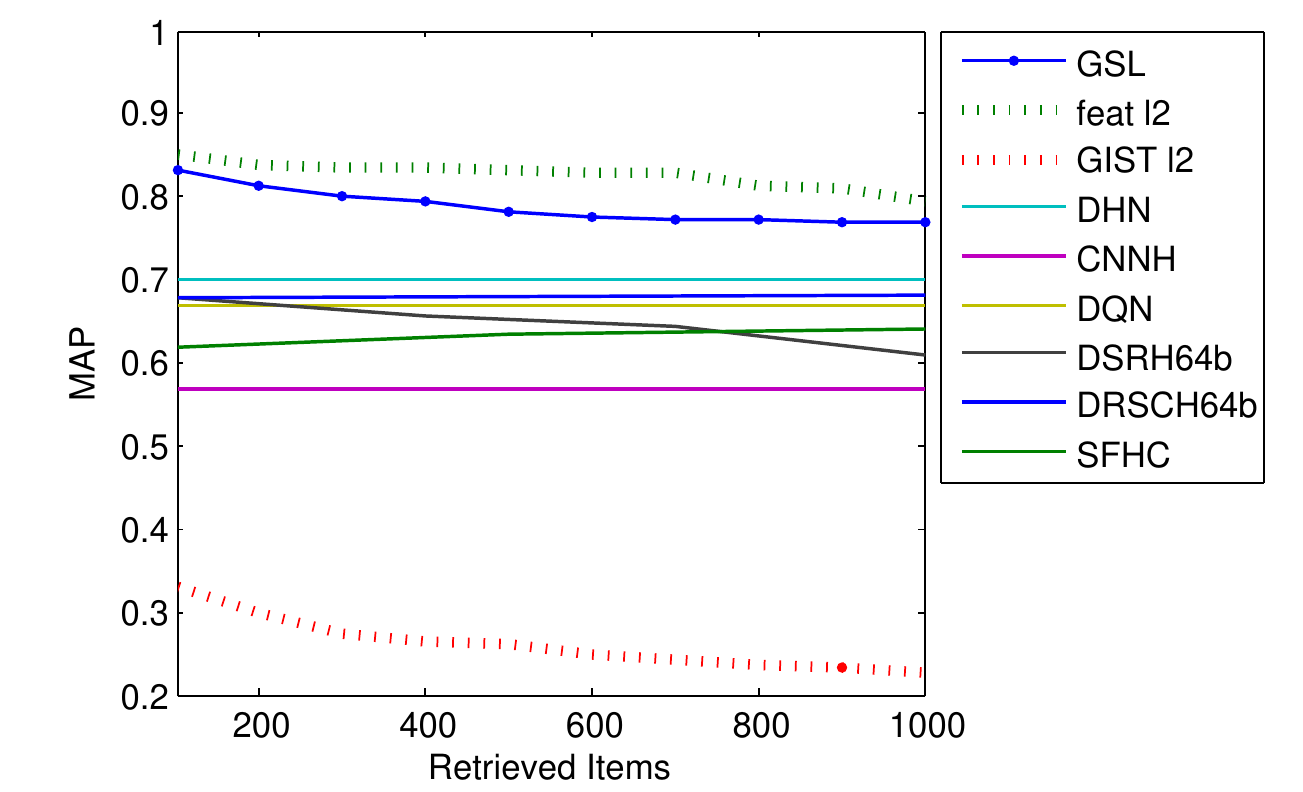}\label{perf:cifar}} 
	\subfigure[NUS-WIDE]{\includegraphics[width=0.340\linewidth]{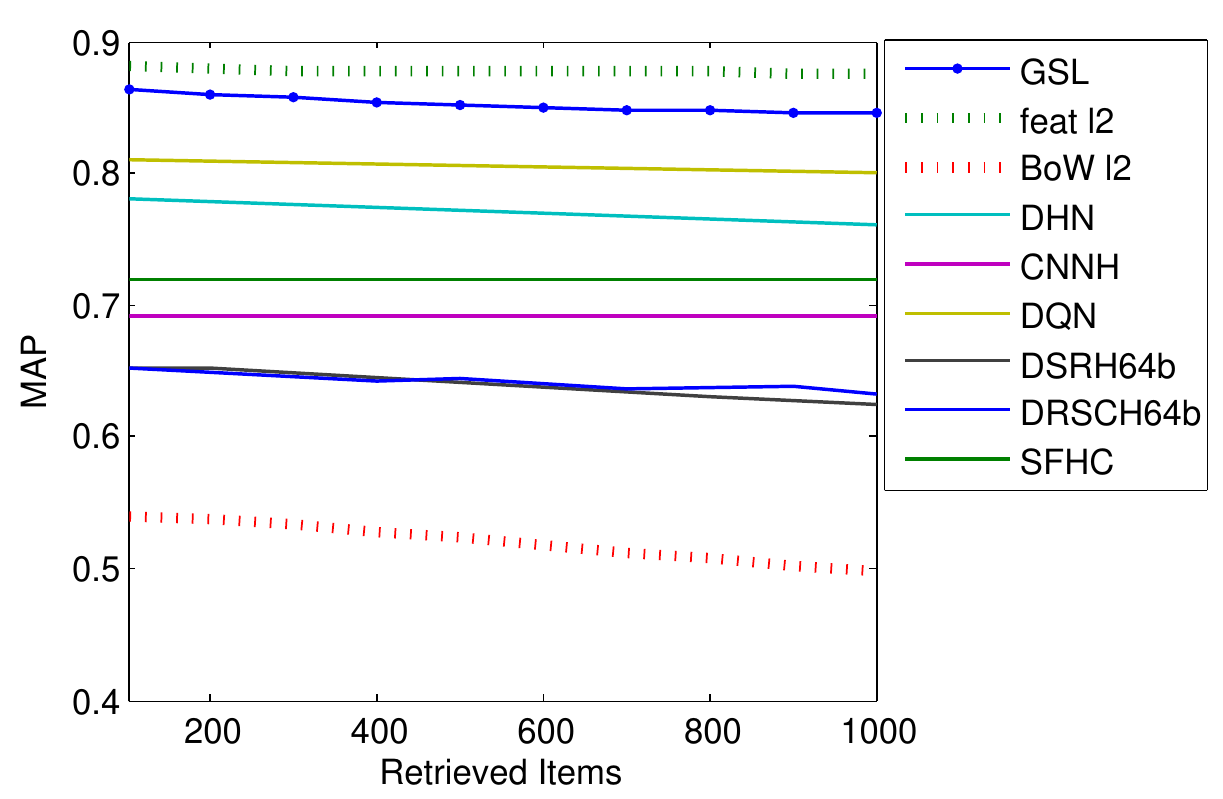}\label{perf:nus}} 
	\subfigure[ILSVRC2012]{\includegraphics[width=0.285\linewidth]{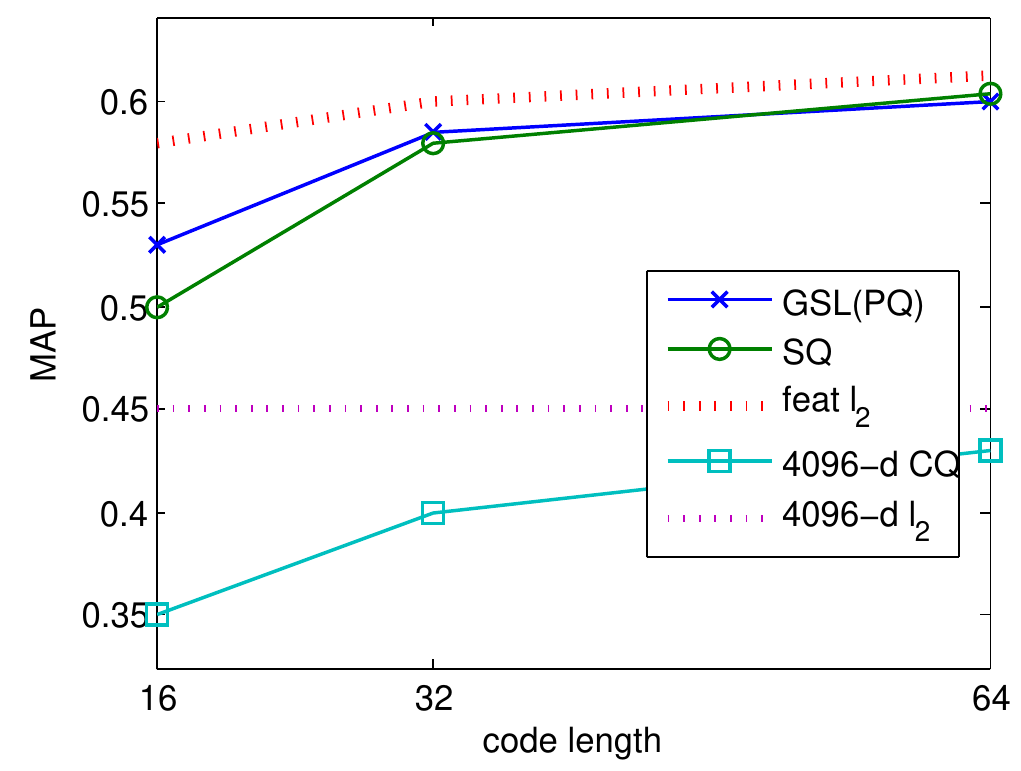}\label{perf:in}} 
	\caption{ Image retrieval performance comparison.
	}
	\label{perf} 
\end{figure*}
\textbf{Regularization effectiveness}: Experiments on CIFAR-10 demonstrate GSL has a significant effect on regularizing the deep representations to be quantized accurately. Specifically, compared to representations without regularization, we observed a 8.9\% performance improvement for PQ and 4.3\% for AQ. This regularization doesn't induce a heavy penalty on the performance of $l_2$ exhaustive search, which can be seen as the performance upper bound for vector quantization methods.

\textbf{Vector quantization methods for GSL}: The choice of different vector quantization method has a small influence on exhaustive $l_2$ search. PQ performs better since it can enumerate neighboring codewords deterministically, so we use PQ for the following experiments. One should always use the same vector quantization method for training and retrieval, which is obvious. Further tests on CIFAR-10 with GSL(PQ)-32bit show that: 1> A higher update frequency of $\mathcal{C}$ improves performances as shown in Fig.\ref{sal:ofq}. 2> Retrieval performances improves when more neighboring codewords are enumerated during training as shown in Fig.\ref{sal:nn}, however it also requires more computation resources.


\textbf{Gradient Snapping vs Output Regularization}: 
Existing simultaneous representation and hash coding learning \cite{zhao2015deep,zhang2015bit,zhu2016deep} or quantization codebook learning \cite{cao2016deep} methods add a penalty term for quantization error for output regularization. This can be seen as a special case of our proposed method when we reduce the number of enumerated neighboring codewords to one, which is then equivalent to adding a constant bias for each sample. In fact, existing vector quantization methods, e.g., \cite{pq,opq,grvq,babenko2014additive} generate codewords combinatorially, and codewords are quite dense in the local space of a representation vector\cite{pq}, thus we can optimize the quantization loss w.r.t. any neighboring codewords. Then the back-propagated gradients can be more conform to the gradients from similarity loss. This makes the learned representations better preserving similarities. We normalize the both gradients and compute the inner-product for each sample on each iteration and report it in Fig.\ref{sal:ovg}, our GSL achieves significantly better conformity between the two gradients.



Note that in extreme cases $(\partial E/\partial \mathbf{y})^T \cdot \Delta \mathbf{c}<0$ for all neighboring codewords, we then set $\lambda_1=\lambda$ and $ \lambda_2=0$ to reject snapping. A better handling of this case might yield better performances.

\subsection{Performance Comparison}
We briefly list the state-of-the-arts simultaneous representation and short code learning methods and the comparison between them in Table\ref{tabcomp}. Since we adopt the same experimental settings, we directly report the experimental result from the corresponding papers. The image retrieval performances measured in MAP on CIFAR10, MNIST, NUS-WIDE are presented in Table \ref{tabr}. Our method out-performs state-of-the-art deep hashing/quantization methods by large margin. We also present retrieval precision w.r.t. top returned samples curves for CIFAR-10 and NUS-WIDE with 48-bit short code obtained with different methods in Fig.\ref{perf:cifar} and Fig.\ref{perf:nus}.

Please note that previous deep representation learning schemes using triplet loss for hashing, e.g.\cite{lai2015simultaneous,zhang2015bit} are outperformed by schemes using pair-wise loss, e.g., \cite{cao2016deep,zhu2016deep}. This doesn't mean triplet loss is inferior to pair-wise loss. In fact, it has been testified that triplet loss works better than pair-wise loss on various tasks including classification\cite{wang2014learning,norouzi2012hamming} and face retrieval\cite{schroff2015facenet}. The main reason is that the quantization regularization introduced in previous triplet loss based method is too weak compared to recent pair-wise loss based methods. By introducing better regularization method, our method achieves significantly better result with triplet loss..

We also compared large scale image retrieval on ILSVRC2012 in Figure \ref{perf:in}. We report the MAP results from \cite{sq} for CQ and SQ. SQ, CQ uses the 4096-d feature extracted from layer \textit{fc7} of AlexNet pretrained on ILSVRC2012. Thus SQ can be seen as adding a "final inner product layer" with output regularization on AlexNet, however SQ doesn't allow the weights of the network to be updated. Our end-to-end representation and short code learning improves the retrieval performances.

\section{Conclusion}
In this paper we propose to regularize gradients on deep convolutional neural network to obtain deep representations exhibiting structures compatible for accurate vector quantization. A gradient snapping layer is discussed detailedly to this end. We validate our approach by training a siamese network architecture based on AlexNet with triplet loss functions to learn representations for similarity search. The experimental results show our approach enforces quantization regularization while learning strong representation, and outperforms the state-of-the-art similarity search methods.

\bibliographystyle{aaai}
\bibliography{nipsbib}

\end{document}